\documentclass{article}
\usepackage{ijcai21}

\usepackage{times}

\usepackage{soul}
\usepackage[hidelinks]{hyperref}
\usepackage[utf8]{inputenc}
\usepackage[small]{caption}
\usepackage{graphicx}
\usepackage{amsmath}
\usepackage{booktabs}
\usepackage{helvet} 
\usepackage{courier}  
\usepackage{amsmath}
\usepackage[switch]{lineno}
\usepackage{color}
\usepackage{booktabs}  
\usepackage{multirow}
\usepackage{threeparttable}  
 
\usepackage{amsthm,amsmath,amssymb}
\usepackage{mathrsfs}
\newcommand{\cut}[1]{}

\usepackage{natbib}  
\usepackage{caption} 
\newcommand{\keypoint}[1]{\vspace{0.1cm}\noindent\textbf{#1}\quad}
\usepackage{amssymb}
\usepackage{pifont}
\newcommand{\cmark}{\ding{51}}%
\newcommand{\xmark}{\ding{55}}%

\title{FedH2L: Federated Learning with Model and Statistical Heterogeneity}
\newcommand{\modelName}{FedH2L}

\author{
Yiying Li$^1$\footnote{liyiying10@nudt.edu.cn}\and
Wei Zhou$^1$\and
Huaimin Wang$^1$\and
Haibo Mi$^1$\and
Timothy M. Hospedales$^{2,3}$\\
\affiliations
$^1$College of Computer, National University of Defense Technology, Changsha, China\\
$^2$School of Informatics, The University of Edinburgh, Edinburgh, UK\\
$^3$Samsung AI Centre, Cambridge\\
}

\begin{document}

\maketitle

\let\thefootnote\relax\footnotetext{This work has been submitted to the IEEE for possible publication. Copyright may be transferred without notice, after which this version may no longer be accessible.}

\begin{abstract}
\begin{quote}
Federated learning (FL) enables distributed participants to collectively learn a strong global model without sacrificing their individual data privacy. Mainstream FL approaches require each participant to share a common network architecture and further assume that data are are sampled IID across participants. However, in real-world deployments participants may require heterogeneous network architectures; and the data distribution is almost certainly non-uniform\cut{ across participants}. To address these issues we introduce \modelName, which is agnostic to \cut{both }the model architecture and robust to different data distributions across participants. In contrast to approaches sharing parameters or gradients, \modelName{} relies on mutual distillation, exchanging only posteriors on a shared seed set between participants in a decentralized manner. This makes it extremely bandwidth efficient, model agnostic, and crucially produces models capable of performing well on the whole data distribution when learning from heterogeneous silos.
\end{quote}
\end{abstract}

\section{Introduction}
Today, artificial intelligence (AI) is showing its strengths in almost every walk of life. To fully realize AI's benefits, we wish to learn models across as much data as possible, but this data is often held privately across diverse users or organizations. To enable collective benefit from AI while maintaining data privacy, Federated Learning (FL) \citep{bonawitz2017secure,google2017fl,Kon2016fl} algorithms aim to train a global model based on the efforts of distributed participants' data and resources. 

There are a number of actively researched challenges however to achieving this vision \citep{li2020federatedChallenges}, including system/model heterogeneity, statistical heterogeneity, bandwidth requirements, and residual privacy concerns. Different FL methods provide different trade-offs in their requirements on these axes along in the accuracy they ultimately provide \citep{li2020federatedChallenges}. We propose a novel FL method \modelName{}, which primarily aims to support significant statistical and model heterogeneity across participants, and also provides benefits for bandwidth and privacy.

System heterogeneity usually refers to different compute and bandwidth resources among participants leading to different update rates among them, and mainstream research aims to alleviate the impact of stragglers in FL setting \citep{li2020fedProx}. However, participants more generally may require fundamentally different model architectures \citep{li2019fedmd}. This can occur in edge or device-based FL due to devices' different memory constraints, or in B2B FL due to each organization wishing to keep their particular optimised model architecture private. Statistical heterogeneity refers to the diversity in each user's data distribution \citep{li2020fedProx,mohri2019afl}. We aim to learn a strong 
federated system capable of performing on the global data distribution, although learning takes place locally in each user's private data silo. 

Mainstream FL methods typically proceed by sharing parameters or gradients at each iteration \citep{google2017fl}. This means they are often bandwidth-constrained, as contemporary models can have millions of parameters. Furthermore, many FL methods require a centralized server to aggregate results from each participant. This requires a globally trusted authority, and provides a single point of failure. In contrast, we present a decentralized peer-to-peer approach that is robust and extremely communication efficient. Parameter and gradient sharing strategies also incur a residual privacy risk due to attack vulnerability \citep{zhu2019deep,luca2018attack}. Our \modelName{} shares no parameters, thus eliminating this vulnerability. 

In this paper, we present a novel FL algorithm \modelName{}, which significantly advances the practical applicability of FL by enabling simultaneous system and statistical heterogeneity across participants. Instead of exchanging gradients/parameters, \modelName{} exchanges predictions on small shared seed set distributed to participants in advance \citep{li2019fedmd}, and performs decentralized global optimization by mutual learning \citep{zhang2018dml}, thus enabling model-agnostic FL. This strategy also eliminates privacy concerns of parameter/gradient sharing, and requires orders of magnitude lower communication cost than sharing models/gradients. However there is still the issue of managing statistical heterogeneity across participants \citep{li2020federatedChallenges,peng2020fada,candela2009dataset}. In \modelName{}, each participant optimizes a multi-task objective of fitting its local data, and distillation on the seed set for knowledge sharing across peers. This multi-task optimization is challenging when there is significant distribution shift\cut{ between participants' data}, which can lead to gradient conflict \citep{yu2020gradient} and poor solutions. To this end we introduce a new optimization strategy to find the best non-conflicting gradient for simultaneously fitting local data and incorporating feedback from peers.
Our contributions are:

\begin{itemize}
\item We introduce \modelName{}, which uniquely provides simultaneous support for a challenging set of real world conditions including \emph{heterogeneous models} across peers, robust \emph{decentralized} learning, \emph{privacy preserving} parameter/gradient-free communication,  while being desired to maximise performance under \emph{heterogeneous data statistics} across peers. \cut{See Table~\ref{tab:tabCompare} for comparison.} 


\item To provide best performance under conditions of heterogeneous data statistics across peers we introduce a new optimization strategy to find the gradient update that does not conflict between local and global update cues.

\item We conduct extensive experiments on several multi-domain datasets: Rotated MNIST \citep{ghifary2015rmnist}, PACS \citep{li2017pacs}, and Office-Home \citep{ven2017office}. Compared to the baselines, we improve the model performance across all domains, demonstrating the effectiveness of \modelName{}. 
\end{itemize}

\begin{table}[]
    \centering
    \resizebox{1.0\columnwidth}{!}{
    \begin{tabular}{rccccc}
    \hline
         Method & Hetero. Models & Decentr. & ParamFree & BW & Hetero. Data \\
         \hline
         FedAvg & \xmark & \xmark & \xmark & - & -\\
         FedProx & \xmark & \xmark & \xmark & - & +\\
         FML & \xmark/\cmark & \xmark & \xmark & - & +\\
         FedGKT & \xmark/\cmark & \xmark & \cmark & + & +\\
         FedMD & \cmark & \xmark & \cmark & + & +\\
         FedAgnostic & \xmark & \xmark & \xmark & - & +\\
         \modelName{} & \cmark & \cmark & \cmark & + & +\\
         \hline
    \end{tabular}}
    \caption{\small Comparison of FL frameworks in terms of support for heterogeneous models, decentralized learning, parameter/gradient-free communication, bandwidth efficiency, and efficacy under settings with heterogeneous data statistics.}
    \label{tab:tabCompare}
\end{table}

\section{Related Work}

\keypoint{System and Statistical Heterogeneity}
FL aims to train models over remote devices, while keeping data localized. FL faces many challenges \citep{li2020federatedChallenges}, and the important one is the heterogeneity on the system and statistical aspects. 

Participants may vary on hardware, compute and bandwidth resources. These system characteristics make issues such as stragglers prevalent. Existing studies mainly focus on the active sampling\cut{ (i.e., actively selecting participants at each round)} \citep{kang2019incentive, nishio2019client}. However, a more severe challenge in system heterogeneity is the model heterogeneity of different  architectures among participants. \citet{li2019fedmd} introduce FedMD for model heterogeneity based on knowledge distillation but with a centralized communication server. FML \citep{shen2020fml} trains extra heterogeneous models by learning from participants' distributed homogeneous models. FedGKT \citep{he2020Group} trains small CNNs on edges and periodically transfer their knowledge (e.g., extracted features) instead of data by knowledge distillation to a server-side large CNN. 

In almost every substantive use case of FL (e.g., medical data across hospitals,\cut{ user data from smartphones,} industrial data across corporations) participants generate and collect data in a \cut{non-identically}Non-IID distributed manner, leading to statistical shift among them. To tackle such statistical heterogeneity, FedProx \citep{li2020fedProx} provides convergence guarantees based on FedAVG \citep{google2017fl}\cut{ when learning} over Non-IID data. FedAgnostic \citep{mohri2019afl} learns a centralized model that is optimized for any target distribution formed by a mixture of participants' distributions. FML \citep{shen2020fml}, FedGKT \citep{he2020Group} and FedMD \citep{li2019fedmd} also have the opportunities to cope with the Non-IID data because they have individualized models for each user but still controlled by a central server.
We aim to handle both model and statistical heterogeneity in a decentralized manner without the need of a centralized model or extra local models.

\keypoint{Bandwidth and Privacy Requirements} 
Communication is a critical bottleneck in FL. The current communication-efficient methods \cut{that try to reduce bandwidth }mainly consider\cut{ two aspects}: (1) Reducing the total number of communication rounds; (2) Reducing the size of transmitted messages at each round. But such methods \citep{google2017fl, li2020fedProx, shen2020fml,mohri2019afl} still typically proceed by sharing the millions of model parameters or gradients as the communicated messages, which means the best case bandwidth requirement is still orders of magnitude worse than \modelName{}. Additionally, sharing  parameters  create attack vulnerability \citep{luca2018attack, zhu2019deep}, increasing privacy risk. The aggregation of parameters and gradients also usually asks for a centralized trusted authority \citep{google2017fl, li2020fedProx} which may lead to the single point of failure. \modelName{} provides a communication-efficient decentralized peer-to-peer method without sharing any high-overhead and privacy compromising model parameters/gradients. 

\keypoint{Multi-task Optimization} 
Instead of learning a single global model, we simultaneously learn distinct local models with a multi-task objective based on local and remote teaching signals. A similar federated work in multi-task setting is MOCHA \citep{smith2017fmtl}, but each local model \cut{in MOCHA }only focuses on the performance on its own task, instead of the multi-task objective. A key challenge in multi-task learning \citep{yu2020gradient, Alex2018Multi} is the conflicting gradients, especially when there is statistical heterogeneity across tasks/participants. \citet{yu2020gradient} propose a gradient surgery \cut{method }to train a single model for multiple tasks by projecting each task gradient onto normal plane of the other. In contrast, we propose a novel optimization strategy to get non-conflicting gradients for each participant's model so as to fit local data and learn from other peers reliably and simultaneously.

\section{Methodology}

Here we introduce the details of \modelName{}. Assume there are $N$ nodes in the FL network, holding data with potentially distinct distributions $\mathcal{D}=\{D_1,D_2,…,D_N\}$. The data on each node contains a set of data-label pairs, i.e., $D_i=\{X_{i}, Y_{i}\}$. We also split $D_i$ into its private data which must only be kept locally, the shared public seed data, validation data and test data, i.e., $D_i=\{D^{\text{pri}}_i,D^{\text{pub}}_i, D^{\text{val}}_i,D^{\text{test}}_i\}$. 
We aim to learn a federated system that aggregates knowledge from all nodes, but without sacrificing each node's data privacy, and without assuming a common model architecture. 
We consider the homogeneous multi-domain setting \citep{li2017pacs}, where all nodes share the same label set $Y_{i}$ covering the same $M$ classes, but have different data distributions. For example, consider medical images of the same set of diseases, but collected by different machines in different hospitals. 
Each node $i$ uses a network parameterized by $\theta_i$,  which can be uniquely customized and private to each node. No centralized model is used in \modelName{}. But the goal is that after learning, each node's model $\theta_i$ should incorporate the knowledge of all nodes' datasets, and be able to perform well on any node's data distribution. 
The workflow is divided into two iterative phases: local and global optimization.



\subsection{Local Optimization}


Local optimization for a node follows the conventional supervised learning paradigm using locally available data.  Denoting  $i$-th node's network as $f_{\theta_{i}}$, we optimize the cross-entropy (CE) loss to obtain gradient $g^\text{loc}_i$:
\begin{equation}\label{eq:theta}
\text{minimize}_{\theta_i}  \ell^{(\text{CE})}(f_{\theta_{i}}(\textbf{x}^{\text{loc}}_{i}), \textbf{y}^{\text{loc}}_{i}),
\end{equation}
\begin{equation}\label{eq:g_pri}
g^{\text{loc}}_i = \nabla_{\theta_{i}} \ell^{(\text{CE})}(f_{\theta_{i}}(\textbf{x}^{\text{loc}}_{i}), \textbf{y}^{\text{loc}}_{i}).
\end{equation}

\cut{
\begin{equation}\label{eq:basic}
\theta_{i} \leftarrow \theta_{i} - \beta \cdot g^{\text{loc}}_i.
\end{equation}
}
Here $d^{\text{loc}}_i=(\textbf{x}^{\text{loc}}_{i}, \textbf{y}^{\text{loc}}_{i})\in \lbrace D^{\text{pri}}_{i},D^{\text{pub}}_{i} \rbrace$ is a batch of the $i$-th domain's data. There is also an alternative setup $d^{\text{loc}}_i=(\textbf{x}^{\text{loc}}_{i}, \textbf{y}^{\text{loc}}_{i})\in \lbrace D^{\text{pri}}_{i},\sum_{n=1}^ND^{\text{pub}}_n \rbrace$ that uses all domains' public seed data. We use the latter option of $d^{\text{loc}}_i$ by default for it behaves slightly better in our experiment, and this is consistent with the data usage strategy in the FL studies with public data \citep{li2019fedmd,zhao2018noniid}. 
Note that  $f_{\theta_{i}}(\textbf{x}^{\text{loc}}_{i})$ provides soft labels $\textbf{p}^{\text{loc}}_i$ corresponding to the output of the final softmax layer of the network, which are compared against the ground truth one-hot labels.
\cut{The local performance is obtained through such conventional supervised learning normally, but MAFML is definitely more than that. Specifically, we hope to challenge the ``stability-plasticity'' dilemma \citep{carpenter1987massively,riemer2019mer}. That is, we pay attention to the ``stability'' (i.e., preserve the performance in the local domain), and also the ``plasticity'' (i.e., generalize to other domains well via mutual learning).}


\subsection{Global Mutual Optimization}
The next step is for each node to learn from its peers. To achieve this in a decentralized manner and under conditions of heterogeneous model architecture, we exploit model distillation. Different from the conventional distillation \citep{hinton2015distill} where a strong teacher trains multiple students, the federated network in \modelName{} acts as an ensemble of students that all teach each other.

\keypoint{Preparation for mutual learning} We  randomly sample a batch $d^{\text{pub}}_i=(\textbf{x}^{\text{pub}}_{i}, \textbf{y}^{\text{pub}}_{i})$ from $D^{\text{pub}}_i$ in each domain/node and compute the soft labels $\textbf{p}^{\text{pub(i)}}_i$. Note that the superscript $i$  denotes the domain the data is drawn from (from the $i$-th domain $d^{\text{pub}}_i$), and the subscript $i$ denotes the network  $f_{\theta_{i}}$ making the prediction. 
To assess the quality of predictions, we also get the accuracy $Acc_i$ over the batch public data in each domain. Each node $i$ broadcasts $[\textbf{p}^{\text{pub(i)}}_i,Acc_i]$ as its teaching signal, and associated teaching confidence, to others in the cohort. Note that the predictions in the teaching signal $\textbf{p}^{\text{pub(i)}}_i$ are with respect to public data $\textbf{x}^\text{pub}_i$, but contain knowledge from the local private data due to being made with the locally optimized network $f_{\theta_i}$. 
The quantities $[\textbf{p}^{\text{pub(i)}}_i,Acc_i]$ are the only parameters exchanged during the federated global mutual optimization step. So this approach is highly communication efficient, and does not disclose any node's private data.


\cut{\keypoint{Node as a teacher.} It is to teach others the current node's domain experience via its domain public data. As $\textbf{p}^{\text{pub(i)}}_i$ and $Acc_i$ have been obtained after the local optimization, it is reasonable to regard $\textbf{p}^{\text{pub(i)}}_i$ as the ``teacher benchmark'' and $Acc_i$ as the ``teaching confidence'' to teach others via $d^{\text{pub}}_i$. }

\keypoint{Mutual Learning} Each node $i$ will act both as a student and a teacher, so there are $(N-1)$ teachers for each student $f_{\theta_i}$. To improve each student node $i$'s model based on teacher node $j$'s data, it is trained to mimic the teacher's soft predictions on the teacher's public data.  Specifically, each student $i$ uses the Kullback Leibler (KL) Divergence loss $\ell^\text{(KL)}_i$ as

\begin{equation}\label{eq:losskl}
\ell^{(\text{KL})}_i = \frac{1}{N-1}\sum_{j=1,j\neq i}^{N} Acc_j * D_{KL}(\textbf{p}^{\text{pub{(j)}}}_{j}||\textbf{p}^{\text{pub{(j)}}}_{i}),
\end{equation}
where each teacher's contribution is weighted by its teaching confidence $Acc_j$, and where 
\begin{equation}\label{eq:kl}
D_{KL}(\textbf{p}^{\text{pub{(j)}}}_{j}||\textbf{p}^{\text{pub{(j)}}}_{i}) = \mathbb{E}_{\textbf{p}_j}[\log\textbf{p}^{\text{pub{(j)}}}_{j} - \log\textbf{p}^{\text{pub{(j)}}}_{i}].
\end{equation}

In addition, besides the KL mimicry loss, we can also take advantage of the conventional supervised loss (CE loss):
\begin{equation}\label{eq:ce_pub}
\ell^{(\text{CE})}_i = \frac{1}{N-1}\sum_{j=1,j\neq i}^{N} \ell^{(\text{CE})}(f_{\theta_{i}}(\textbf{x}^{\text{pub}}_{j}), \textbf{y}^{\text{pub}}_{j}),
\end{equation}

Thus we obtain the total mutual learning gradient for node $i$ learning from the other nodes in the cohort:
\begin{equation}\label{eq:g_glo}
g^{\text{pub}}_i = \nabla_{\theta_{i}} (\ell^{(\text{KL})}_i+\ell^{(\text{CE})}_i).
\end{equation}

\keypoint{Summary} In summary, each node trains using $g^\text{loc}_i=\nabla_{\theta_i}\ell^\text{(CE)}$  on local data, and $g^\text{pub}_i=\nabla_{\theta_i}(\ell^\text{(KL)}_i+\ell^\text{(CE)}_i)$ using other domains' public seed data.

\cut{So the nodes in MAFML act as ``student-teacher'' cohorts.
Now based on $\ell^{(\text{CE})}$ on locally accessible data, we obtain $g^{\text{loc}}_{i}$ and indeed update $\theta_i$ using $g^{\text{loc}}_{i}$ during the local optimization; and based on $\ell^{(\text{KL})}+\ell^{(\text{CE})}$ on the domains' public data, we can obtain $g^{\text{pub}}_{i}$. So the question we need to think about at present is: will we then update $\theta_i$ by directly using $g^{\text{pub}}_{i}$ in the global optimization stage?}

\subsection{Dealing with Statistical Heterogeneity} Our algorithm described so far enables decentralized FL of heterogeneous \emph{models}. However, a key challenge is to best support the practically ubiquitous situation of \emph{statistical} heterogeneity across domains. We hope that the local gradient $g^\text{loc}_i$ can help to improve the performance on other domain's data (Cross-Domain Performance), and the remote teacher gradient $g^\text{pub}_i$ lcan help to improve the performance on the local data (Within-Domain Performance). However this is challenging to achieve from a multi-task learning perspective, because the local learning gradient and peer learning gradient may conflict \citep{yu2020gradient,paz2017gem,zhou2021gradient}  under significant statistical shift. 


\keypoint{Mutual Learning robust to statistical shift} To perform student-teacher learning that is robust to distribution-shift across nodes, we propose to enforce the constraint:

\begin{equation}
\left\langle g^{\text{loc}}_i, g^{\text{pub}}_i \right\rangle \ge 0.
\label{eq:constraints}
\end{equation}

If this constraint is satisfied, then the remote teaching signal  $g^{\text{pub}}_{i}$ is unlikely to increase $\ell^{(\text{CE})}$ on each domain's local data, and we can safely use $g^{\text{pub}}_{i}$ to directly update $\theta_i$ without risking negative within-domain performance. Thus we check if the constraint is violated, and project $g^{\text{pub}}_{i}$ to the closest gradient $\tilde{g}_{i}$ (in the $\ell_2$ norm sense) satisfying constraint (\ref{eq:constraints}). After projection $\tilde{g}_{i}$ is unlikely to increase $\ell^{(\text{CE})}$ or $\ell^{(\text{KL})}$. We perform:

\begin{align}
    \text{minimize}_{\tilde{g}_{i}} \quad&\frac{1}{2} \|g^{\text{pub}}_i - \tilde{g}_{i}\|_2^2\nonumber\\ 
    \text{subject to} \quad& \langle \tilde{g}_{i}, g^{\text{loc}}_i \rangle \ge 0, \mbox{ for all } i \in N.\label{eq:gemprimal}
\end{align}

\keypoint{Computation of $\tilde{g}_{i}$} We set $\tilde{g}_i \leftarrow project(g^{\text{pub}}_i, g^{\text{loc}}_i)$. Here\cut{ the function} $project$ is the optimization of dual problem of Quadratic Program (QP). To solve \eqref{eq:gemprimal} efficiently, recall the primal of a QP \citep{nocedal2006numerical} with inequality constraints:
\begin{align}
    \text{minimize}_z \quad&\frac{1}{2} z^\top C z + w^\top z\nonumber\\
    \text{subject to} \quad&Az \leq b,\label{eq:primal}
\end{align}
where $C \in \mathbb{R}^{p \times p}$ is a real symmetric matrix, $w\in \mathbb{R}^p$ is a real-valued vector , $A^\top \in
\mathbb{R}^{p}$ is a real matrix, and $b\in\mathbb{R}$, $p$ is the dimension of gradient vector.

The solution to the dual problem provides a lower bound to \cut{the solution of }the primal QP problem. The Lagrangian dual of a QP is also a QP. Because original problem has constraint conditions, these can be built into the function. We write the Lagrangian function \citep{bot2009duality} as:
\begin{equation}
L(z,v )=\frac{1}{2} z^\top C z +w^\top z+ v ^\top (Az-b).
\end{equation}

Defining the (Lagrangian) dual function as $g(v)=\inf _{z}L(z,v )$, we find an infimum of $L$, which occurs where the gradient is equal to zero, using $\nabla _{z}L(z,v )=0$ and positive-definiteness of Q:

\begin{equation}
z^{*}=-C^{-1}(A^\top v+w).
\end{equation}

So, the dual problem of~\eqref{eq:primal} 
is:
\begin{align}
    \text{minimize}_{v}   \quad&\frac{1}{2} v^\top AC^{-1}A^\top v + (w^\top C^{-1}A^\top + b^\top) v\nonumber\\
    \text{subject to}   \quad&v \geq 0.\label{eq:dual}
\end{align}

With these notations, we write the primal QP~\eqref{eq:gemprimal} as:
\begin{equation}
    \begin{split}
    \text{minimize}_z  \quad& \frac{1}{2} z^\top z -{g^{\text{pub}}_i}^\top z + \frac{1}{2} {g^{\text{pub}}_i}^\top {g^{\text{pub}}_i}\\
    \text{subject to}   \quad& -{g^{\text{loc}}_i}^\top z \leq 0.\label{eq:qp_org}
    \end{split}
\end{equation}

According to the conversion formula above, We can pose the dual of the \modelName{} QP as:

\begin{align}
    \text{minimize}_{v}   \quad&\frac{1}{2} v^\top {g^{\text{loc}}_i}^\top {g^{\text{loc}}_i} v + {g^{\text{pub}}_i}^\top {g^{\text{loc}}_i} v\nonumber\\
    \text{subject to} \quad&v \geq 0.
    \label{eq:project}
\end{align}

After \eqref{eq:project} is solved for $v^\star$ which is specifically a real number here, we reset the projected gradient as $\tilde{g}_i = v^\star {g^{\text{loc}}_i}  + g^{\text{pub}}_i$, and use $\tilde{g}_i$ to update $\theta_i$ for the global mutual optimization.

\subsection{Summary}

\cut{Bringing all components together, we have the full algorithm in Algo.~\ref{alg:main}. }To summarize, (1) in each domain/node we first perform a local update with $g^{\text{loc}}_i$ using $\ell^{(\text{CE})}$ on the locally preserved data and then broadcast its teaching signal $[\textbf{p}^{\text{pub(i)}}_i, Acc_i]$ on its public data. (2) In the global mutual optimization, \modelName{} introduces distillation mimicry loss $\ell^{(\text{KL})}$ in addition to the conventional $\ell^{(\text{CE})}$ in order for each node to learn from its peers' teaching signals. (3) To manage potential conflicting gradients across nodes between $g^{\text{loc}}_i$ and $g^{\text{pub}}_i$, we calculate the projected gradient $\tilde{g}_i$ as the final global gradient to update each $f_{\theta_i}$. This ensures that each node in the cohort achieves both CDP and WDP, improving performance on its own data, as well as strengthening its model to perform well on the private statistically heterogeneous distributions held by other nodes.  This is the first work to consider both model and statistical heterogeneity across nodes in FL.

\section{Experiments}
We evaluate \cut{our approach }on 
digit classification (Rotated MNIST) and image recognition (PACS, Office-Home) tasks.
These datasets contain multiple sub-domains with statistical shift. We use Ray \citep{moritz2018ray} framework to implement distributed applications\cut{ on different nodes}. \cut{Experiments are implemented by PyTorch and run on a server with 4 GPUs. }We compare \modelName{} to the alternatives:

\begin{itemize}
  \item {\em Independent (IND):} Node only uses its own domain (pri+pub) data for conventional training (SGD on CE).
  \item \emph{Aggregation (AGG):} Node aggregates its own (pri+pub) data and the shared public data from other nodes for conventional training.  AGG is usually a strong baseline to beat in multi-domain learning  \citep{li2019fc}.
  \item \emph{FedMD \citep{li2019fedmd}:} A state of the art centralized approach to model-heterogenity in FL.
  \item \emph{FedAvg \citep{google2017fl}:} The classic FL method that uses a central server to aggregate gradients and distribute parameters.
  \item \emph{FedProx \citep{li2020fedProx}:} A FedAvg-based approach that provides convergence guarantees for learning over statistical heterogeneity.
\end{itemize}

\keypoint{Metrics} \cut{In centralized approaches to FL, a single model is ultimately learned. }In our decentralized approach, each node has its own model, and our goal is all models should outperform that of a centralized competitor such as FedAvg. So we report the average test performance across all nodes' models. Considering the statistical heterogeneity, we report the following three metrics, where $F$ evalautes test accuracy\cut{ on the test data, and the subscript $i$ of $F$ denotes using the model of $i$-th node and the data in the brackets is the applied test data}. 

\noindent\underline{\textit{Within-Domain Performance:}} $WDP_i = F_i(D^{\text{test}}_i)$. WDP is the performance of $f_{\theta_{i}}$ on the node $i$'s test data. Higher WDP values indicate the learning experience from other nodes improve the performance on the current node. This is not guaranteed by a simple FL algorithm as other nodes' gradients can potentially cause conflict  or forgetting \citep{yu2020gradient,mccloskey1989cata}. \modelName{} aims to improve WDP by projecting away conflicting gradients. 

\cut{
\begin{equation}
WDP_i = F_i(D^{\text{test}}_i).
\label{eq:bwt}
\end{equation}
}

\noindent\underline{\textit{Cross-Domain Performance:}} $CDP_i = F_i(\sum_{n=1,n\neq i}^ND^{\text{test}}_n)$. CDP is the performance of $f_{\theta_{i}}$ on all other nodes' test data. If FL nodes do not learn from their peers then CDP will be low due to statistical shift.

\cut{
\begin{equation}
CDP_i = F_i(\sum_{n=1,n\neq i}^ND^{\text{test}}_n).
\label{eq:fwt}
\end{equation}
}

\noindent\underline{\textit{Average accuracy:}} $ACC_i = F_i(\sum_{n=1}^ND^{\text{test}}_n)$. ACC is the all-domain performance of $f_{\theta_{i}}$ on all nodes' test data.

\cut{
\begin{equation}
ACC_i = F_i(\sum_{n=1}^ND^{\text{test}}_n).
\label{eq:acc}
\end{equation}
}

\subsection{Evaluation on Rotated MNIST}
\keypoint{Dataset and settings}  Rotated MNIST \citep{ghifary2015rmnist} contains different domains with each one corresponding to a degree of roll rotation in \cut{the classic} MNIST dataset. \cut{So different node has the different data statistics. We take this idea to build the Rotated MNIST dataset for our experiment.} The basic view (M0) is formed by randomly choosing 100 images each of ten classes from \cut{the original} MNIST dataset, and we \cut{then} create 3 rotating domains from M0 with $20^\circ$ rotation each in clockwise direction, denoted M20, M40, M60.\cut{, and the image dimensions remain unchanged.} The data on each node is split by default $65\%/10\%/10\%/15\%$ for  $D^{\text{pri}}_i/D^{\text{pub}}_i/ D^{\text{val}}_i/D^{\text{test}}_i$.

We first experiment by easily deploying homogeneous networks (e.g. LeNet \citep{lecun1998lenet})\cut{ on these nodes}. We train \cut{these methods }using AMSGrad \citep{reddi2018adam} optimizer (lr=1e-3, weight decay=1e-4) for 10,000 rounds and set batch\_size=32. We explore performance considering several factors: (1) $\alpha$, the proportion of \cut{shared public data }$D^{\text{pub}}_i$. We set the proportion of $(D^{\text{pri}}_i+D^{\text{pub}}_i)$ as $75\%$, and $D^{\text{val}}_i$ and $D^{\text{test}}_i$ account for $10\%$ and $15\%$ unchanged respectively. Note that the performance of IND, FedAvg and FedProx is independent of \cut{the value of} $\alpha$. (2) In \modelName{}, $E$ is the ratio between global and local update rounds. Local optimization is carried out each round, and global optimization every $E$ rounds. 
So  when calculating the global update $\tilde{g}_i$, $g^{\text{loc}}_i$ is actually $(g^{\text{loc}}_{i\_E} - g^{\text{loc}}_{i\_0})$ over $E$ rounds. Here we set default $E=1$, and then ablate the hyperparameter sensitivity on $E$. 
(3) We explore both homogeneous and heterogeneous architectures. Note that even in the homogeneous architecture case, decentralized \modelName{} nodes have independent parameters.
\cut{Here note that even if they are in the same architecture, their network parameters are different since each node updates its model independently without the share of parameters or gradients. When we mention the network personality, we are referring to the architecture and network parameters.}


\begin{table*}[htb]
  \centering  
  \fontsize{7.4}{7}\selectfont  
  \begin{threeparttable}  
  \caption{Test result (\%) on three metrics on Rotated MNIST.}  
  \label{tab:performance_comparison_mnist}  
    \begin{tabular}{lccccccccccccccc}  
    \toprule  
    \multirow{2}*{Method}&  
    \multicolumn{3}{c}{ M0-LeNet}&\multicolumn{3}{c}{ M20-LeNet}&\multicolumn{3}{c}{ M40-LeNet}&\multicolumn{3}{c}{ M60-LeNet}&\multicolumn{3}{c}{ Avg.}\cr  
    \cmidrule(lr){2-4} \cmidrule(lr){5-7} \cmidrule(lr){8-10}\cmidrule(lr){11-13}\cmidrule(lr){14-16}  
    &ACC&WDP&CDP&ACC&WDP&CDP&ACC&WDP&CDP&ACC&WDP&CDP&ACC&WDP&CDP\cr
    \midrule 
    \modelName{} ($\alpha$=5\%)&{\bf86.17}&88.67&{\bf85.33}&86.33&{\bf93.33}&85.11&{\bf87.50}&93.33&{\bf 85.78}&{\bf87.17}&{\bf96.00}&{\bf84.22}&{\bf86.79}&{\bf92.83}&{\bf85.11}\cr

    AGG ($\alpha$=5\%)&85.50&{\bf92.67}&83.11&{\bf87.50}&{\bf93.33}&{\bf85.56}&83.67&90.00&81.56&83.83& 93.33&80.67&85.13&92.33&82.73\cr
    FedMD ($\alpha$=5\%) &84.17&87.33&83.11&85.33&91.33&83.33&86.67&{\bf96.00}&83.56&84.17&91.33&81.78&85.09&89.11&82.95\cr

    \midrule  
    \modelName{} ($\alpha$=10\%)&90.17&{\bf 93.33}& 89.11&{\bf 91.67}&{\bf 96.00}&{\bf 90.22}&86.50&90.67&{\bf85.11}&{\bf 88.17}&{\bf 93.33}& {\bf86.44}&{\bf89.13}&{\bf93.33}&{\bf87.72}\cr  
    
    AGG ($\alpha$=10\%)&86.50&90.00&85.33&87.17&92.67&85.33&{\bf86.67}&{\bf94.00}&84.22&80.67&91.33&77.11&85.25&92.00&83.00\cr 
    FedMD ($\alpha$=10\%) &85.00&88.67&83.78&87.67&95.33&85.11&82.00&90.67&79.11&85.67&90.00&84.22&85.09&91.17&83.06\cr
    
    \modelName{} (asynchronous)&{\bf90.66}&{\bf93.33}&{\bf89.78}&90.00&94.00&88.67&85.50&90.67&83.78&86.67&90.67&85.33&88.21&92.17&86.89\cr

    \midrule 
    \modelName{} ($\alpha$=15\%)&{\bf89.67}&91.33&89.11&{\bf90.00}&92.67&{\bf89.11}&{\bf90.50}&{\bf94.00}&{\bf89.33}&{\bf88.33}&{\bf92.67}&86.89&{\bf89.63}&{\bf92.67}&{\bf88.61}\cr

    AGG ($\alpha$=15\%)&87.83&{\bf92.00}&86.44&89.67&92.10&88.44&87.83&{\bf94.00}&85.78&86.00&91.33& 84.22&87.83&92.47&86.22\cr
    FedMD ($\alpha$=15\%) &88.67&89.33&88.44&89.00&93.33&87.56&85.00&90.00&83.33&84.33&{\bf92.67}&81.56&86.75&91.33&85.22 \cr
    
    IND &66.39&91.33&58.08&78.11&{\bf94.00}&72.82&72.39&93.11&65.48&56.89&91.78&45.48&68.45&92.56&60.47\cr
    
    FedAvg &86.50&77.33&{\bf89.56}&86.50&86.67&86.44&86.50&92.67&84.44&86.50&89.33&85.56&86.50&86.50&86.50\cr
    
    FedProx &86.67&80.00&88.89&86.67&90.00&85.56&86.67&91.33&85.11&86.67&85.33&{\bf87.11}&86.67&86.67&86.67\cr

    \bottomrule  
    \end{tabular}  
    \end{threeparttable}  
\end{table*} 

\keypoint{Results} Table~\ref{tab:performance_comparison_mnist}
shows the results including varying \cut{hyperparameters} $\alpha$ \cut{and $E$} of \modelName{}. We evaluate using the validation data every 50 rounds and keep the model with the maximal ACC for the final test on three metrics. Max value on each metric is bold. \cut{From the results,} We \cut{can} draw the following conclusions: 
(1) \modelName{} generally outperforms  competitors for a range of $\alpha$\cut{ and $E$}.
(2) \modelName{} generally performs better with increased public data proportion $\alpha$\cut{ and lower update interval $E$. Performance degrades smoothly with larger $E$ which lowers communication cost proportionally}.
(3) \modelName{} outperforms the AGG and IND  baselines at every $\alpha$ operating point.
(4) Compared to state of the art competitors, \modelName{} outperforms FedMD at every operating point. The poor performance of FedMD compared to \modelName{} and AGG shows that it is vulnerable to distribution shift between domains. The vanilla centralized FedAVG/FedProx require over 1000$\times$ the communication bandwidth of \modelName{}, and we now restrict their bandwidth to match that used by \modelName{} and get the results in Table~\ref{tab:performance_comparison_mnist}. 
\modelName{} outperforms FedAvg/FedProx clearly at $\alpha=15\%$.

\begin{figure}[tb]
\centering
\includegraphics[width=0.15\textwidth]{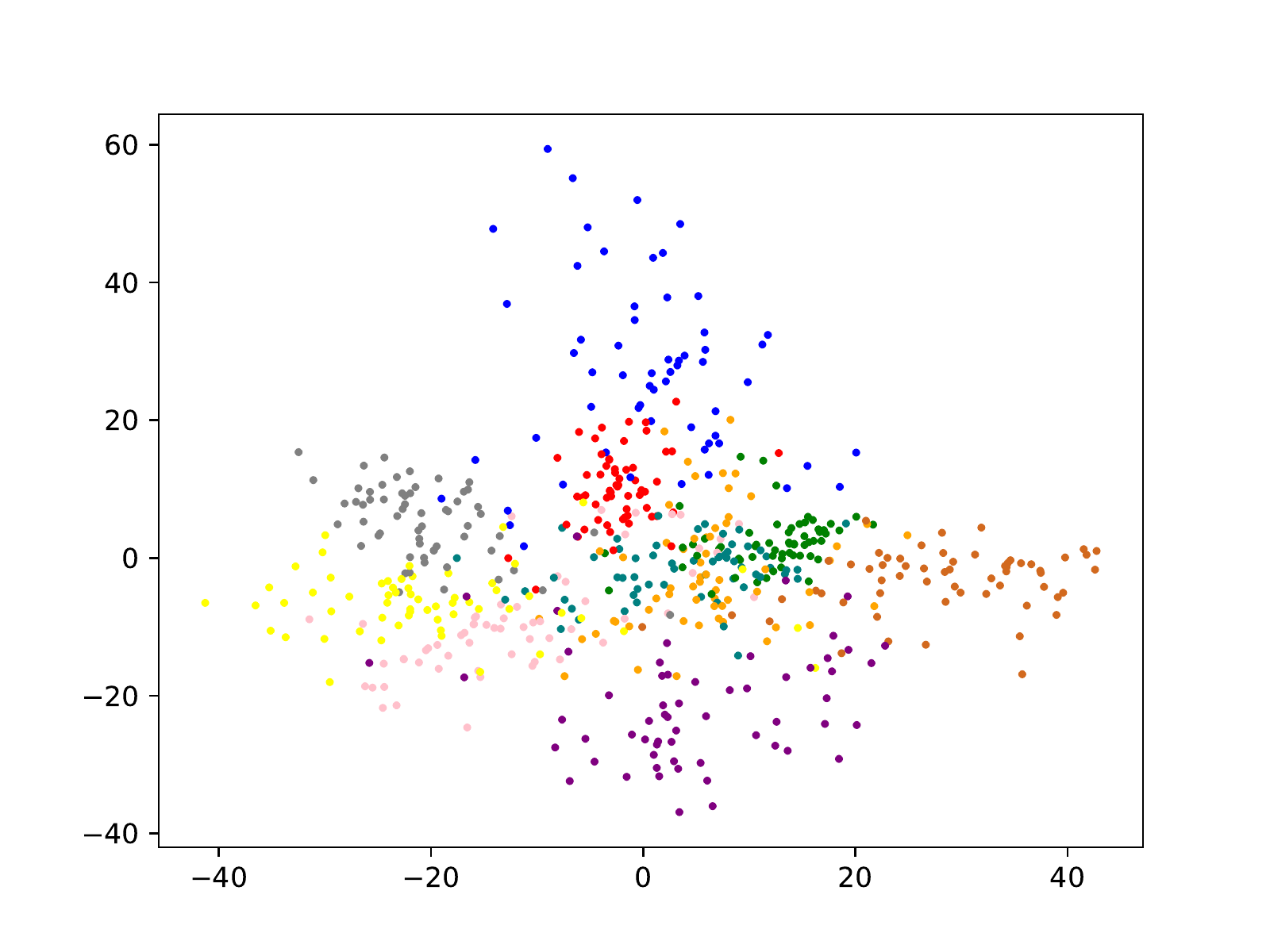}
\includegraphics[width=0.15\textwidth]{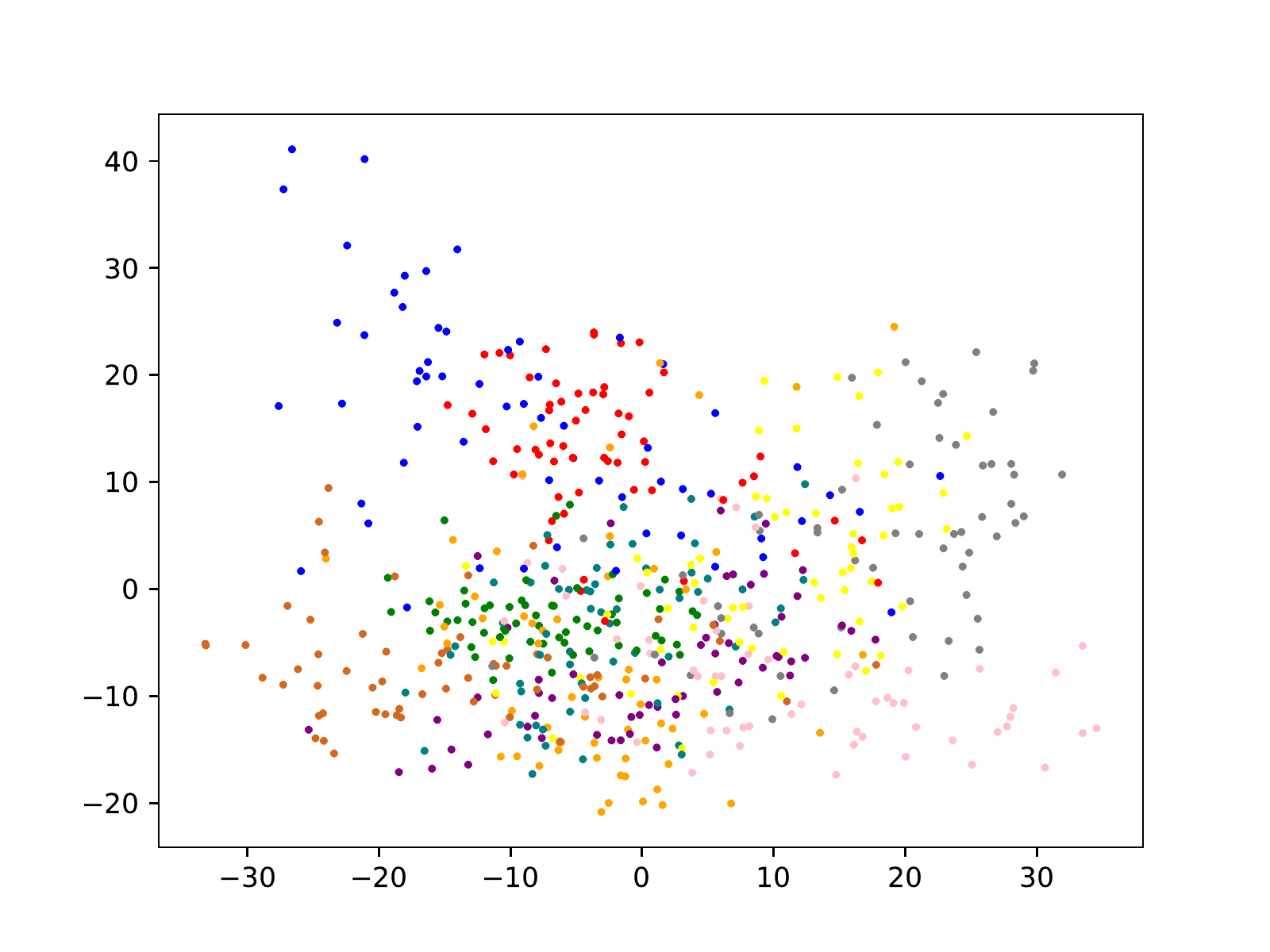}
\includegraphics[width=0.15\textwidth]{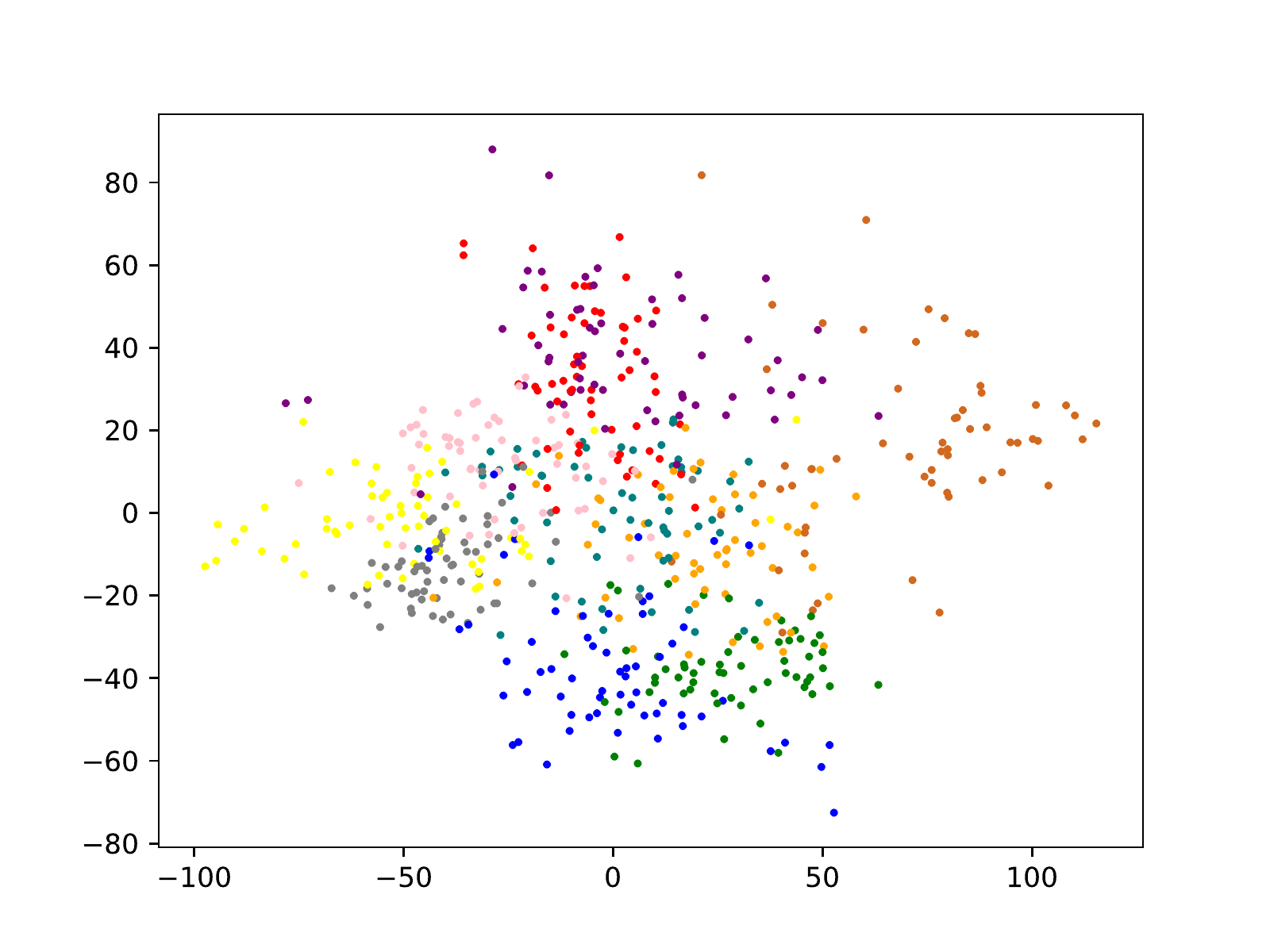}
\caption{PCA projections of features on all domains' test data using domain M0's  model of Rotated MNIST for example. Left: \modelName{}. Middle: IND. Right: AGG. (Dot: Image. Color: Digit label.)}
\label{fig:pca}
\end{figure}

\keypoint{Qualitative Results}
\cut{To qualitatively visualize the results, }We perform PCA projections of the features on all domains' test data in Figure~\ref{fig:pca}.
\modelName{} provides the improved overall separability on all domains' data.

\begin{table*}[htbp]
  \centering  
  \fontsize{7.4}{7}\selectfont  
  \begin{threeparttable}  
  \caption{Test result (\%) on three metrics on PACS with heterogeneous models.}  
  \label{tab:performance_comparison_pacs}  
    \begin{tabular}{lccccccccccccccc}  
    \toprule 
    \multirow{2}*{Method}&  
    \multicolumn{3}{c}{ Photo-ResNet18}&\multicolumn{3}{c}{ Art\_painting-ResNet34}&\multicolumn{3}{c}{ Cartoon-AlexNet}&\multicolumn{3}{c}{ Sketch-VGG11}&\multicolumn{3}{c}{ Avg.}\cr  
    \cmidrule(lr){2-4} \cmidrule(lr){5-7} \cmidrule(lr){8-10}\cmidrule(lr){11-13}\cmidrule(lr){14-16}  
    &ACC&WDP&CDP&ACC&WDP&CDP&ACC&WDP&CDP&ACC&WDP&CDP&ACC&WDP&CDP\cr  
    \midrule  
    \modelName{} & 83.86&99.80&80.66&{\bf 90.91}&99.95&{\bf 88.57}&{\bf81.68}&{\bf99.67}&{\bf76.16}&{\bf52.87}&{\bf80.33}&{\bf37.26}&{\bf77.33}&{\bf94.94}&{\bf70.66}\cr  
    IND &51.08&99.57&41.29&77.72&99.30&72.15&68.52&99.39&59.05&44.79&78.75&22.83&60.53&94.25&48.83\cr
    AGG &{\bf84.90}&{\bf 100.00}&{\bf81.90}&89.50&{\bf 100.00}&86.85&80.80&98.77&75.28&52.81&78.01&36.51&77.00&94.20&70.14\cr
    FedMD &80.05&{\bf100.00}&76.05&86.90&99.08&83.75&78.07&95.65&72.67&51.40&75.47&35.83&74.11&92.55&67.08\cr
    FedAvg/FedProx &-&-&-&-&-&-&-&-&-&-&-&-&-&-&-\cr
    \bottomrule  
    \end{tabular}  
    \end{threeparttable}  
\end{table*}

\subsection{Evaluation on PACS dataset}
\keypoint{Dataset and settings} PACS \citep{li2017pacs} is a multi-domain object recognition benchmark with  9991 images of  7 categories across 4 different domains.
The original PACS dataset has a fixed split for train, validation and test. We separate out 10\% of its test part as the public seed data, and directly use the train part as our private data.
Here we mainly consider the heterogeneous model case where we randomly deploy ResNet18, ResNet34, AlexNet and VGG11\cut{ as their networks to each node}. The homogenous model case where all nodes use a ResNet18 is reported in the supplementary material, and it also shows the benefits of \modelName{}. We use AMSGrad (lr=1e-4, weight decay=1e-5) to train 10,000 rounds and set batch\_size=32. 

\keypoint{Results} We can see from Table~\ref{tab:performance_comparison_pacs}: \cut{(i) \modelName{} generally provides a consistent improvement over other methods in the homogeneous case (top). The original communication bandwidth of FedAvg is $\sim$10e6 times that of ours when using ResNet18. Even if we control FedAvg's communication to $\sim$100 times to ours by controlling $E$ and its participating fraction \citep{google2017fl}, and then get the results in Table~\ref{tab:performance_comparison_pacs}, FedAvg is still outperformed by \modelName{}.} (i) In the heterogeneous case, FedAvg and FedProx are inherently inapplicable and \modelName{} surpasses the other alternatives. (ii) We observe that although VGG11 does not perform well in the sketch domain (see IND/AGG WDP), when used with \modelName{}, it still benefits rather than harms the other nodes' performance thanks in part due to the teaching confidence signal (Eq.~(\ref{eq:losskl})). 

\subsection{Evaluation on Office-Home dataset}
\keypoint{Dataset and settings} The Office-Home \citep{venkateswara2017Deep} dataset is initially proposed to evaluate domain adaptation\cut{ for object recognition}. It consists 4 different domains\cut{: Artistic, Clip Art, Product and Real-World}\cut{ images} with each \cut{domain }containing images of 65 object categories. We split each domains data into $\{D^{\text{pri}}_i,D^{\text{pub}}_i, D^{\text{val}}_i,D^{\text{test}}_i\}$ according to the default ${[65\%,10\%,10\%,15\%]}$. We randomly apply ResNet34, MobileNet, AlexNet and ResNet50 as their heterogeneous models and  use the same hyperparameters as in the PACS experiment. The homogeneous model case is also reported in the supplementary material where \modelName{} shows consistent benefits.

\begin{table*}[h!]
  \centering  
  \fontsize{7.4}{7}\selectfont  
  \begin{threeparttable}  
  \caption{Test result (\%) on three metrics on Office-Home with heterogeneous models.}  
  \label{tab:performance_comparison_office}  
    \begin{tabular}{lccccccccccccccc}  
    \toprule
    \multirow{2}*{Method}&  
    \multicolumn{3}{c}{ Art-ResNet34}&\multicolumn{3}{c}{ Clipart-MobileNet}&\multicolumn{3}{c}{ Product-AlexNet}&\multicolumn{3}{c}{ Real\_world-ResNet50}&\multicolumn{3}{c}{ Avg.}\cr  
    \cmidrule(lr){2-4} \cmidrule(lr){5-7} \cmidrule(lr){8-10}\cmidrule(lr){11-13}\cmidrule(lr){14-16}  
    &ACC&WDP&CDP&ACC&WDP&CDP&ACC&WDP&CDP&ACC&WDP&CDP&ACC&WDP&CDP\cr 
    \midrule
    \modelName{} &{\bf 65.52}&{\bf58.70}&{\bf 66.70}&{\bf 73.55}&76.52&{\bf 72.40}&{\bf 59.64}&{\bf80.82}&{\bf 51.00}&{\bf 60.97}&{\bf70.29}&{\bf 57.30}&{\bf64.92}&{\bf71.58}&{\bf61.85}\cr  
    IND &41.00&57.14&38.20&55.14&{\bf78.49}&46.08&46.60&79.40&33.23&47.61&63.31&41.42&47.59&69.59&39.73\cr
    AGG &57.34&51.86&58.30&70.61&{\bf78.49}&67.56&54.32&77.02&45.05&54.68&64.94&50.64&59.24&68.08&55.39\cr
    FedMD &55.46&55.59&55.44&67.49&77.50&63.61&53.17&75.59&44.02&51.74&59.42&48.72&56.97&67.03&52.95\cr
    FedAvg/FedProx &-&-&-&-&-&-&-&-&-&-&-&-&-&-&-\cr
    \bottomrule  
    \end{tabular}  
    \end{threeparttable}  
\end{table*} 

\keypoint{Results}
In Table~\ref{tab:performance_comparison_office}, \modelName{} gives a clear boost to overall accuracy, within-domain and cross-domain performance\cut{ for both heterogeneous and homogeneous settings}. 

\subsection{Further Analysis}

\begin{figure}[tb]
\centering
\includegraphics[width=0.15\textwidth]{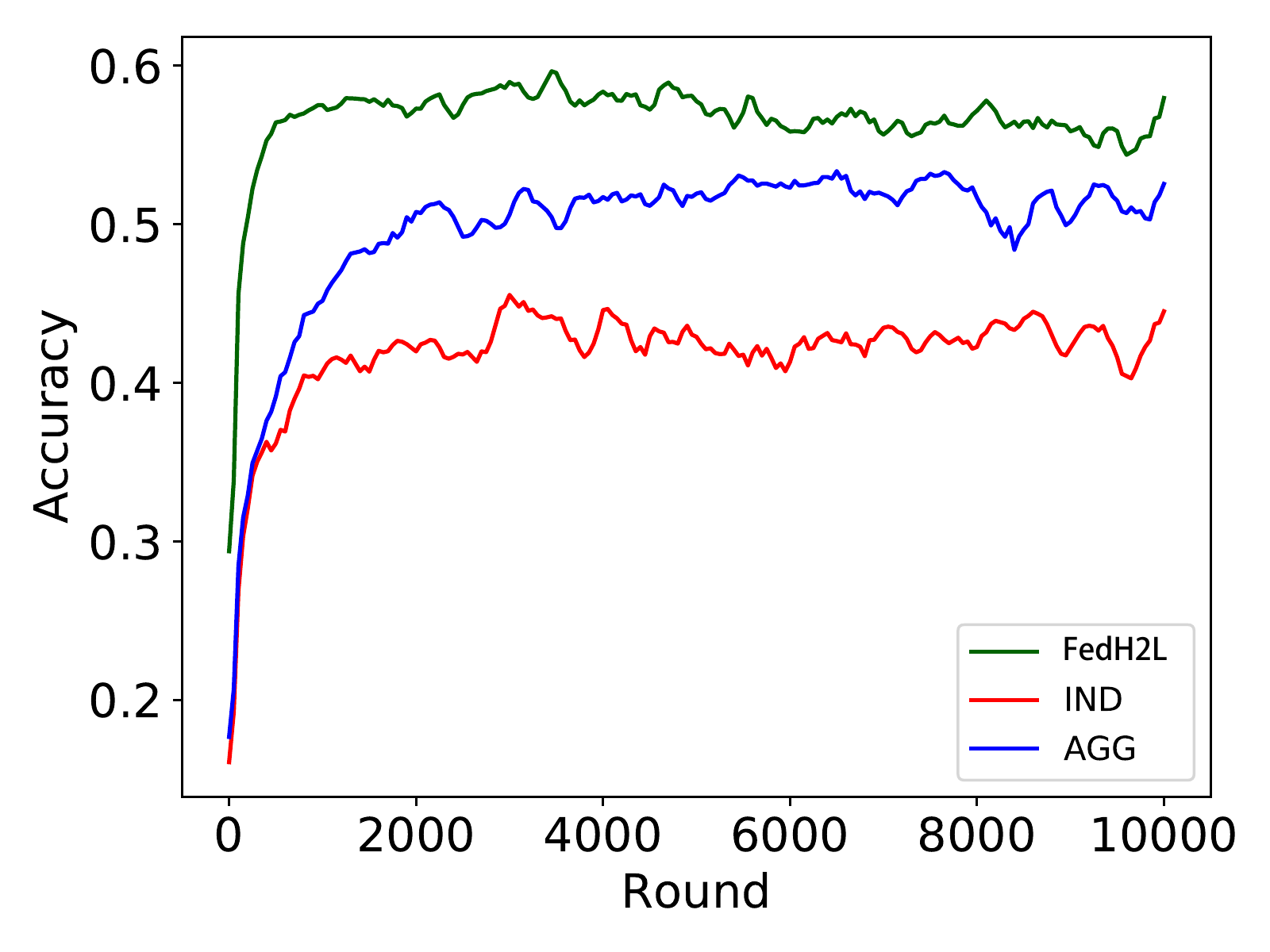}
\includegraphics[width=0.15\textwidth]{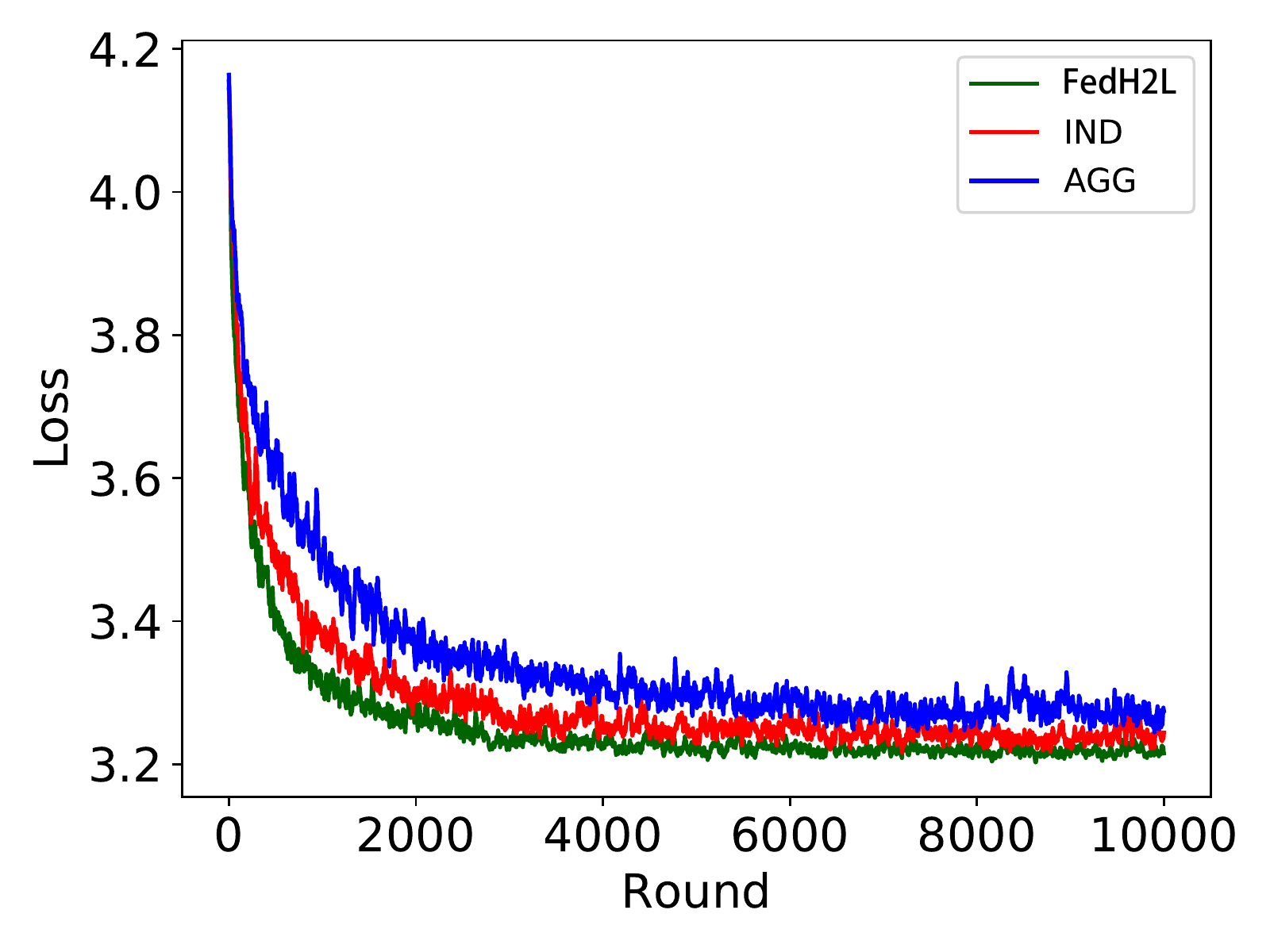}
\includegraphics[width=0.15\textwidth]{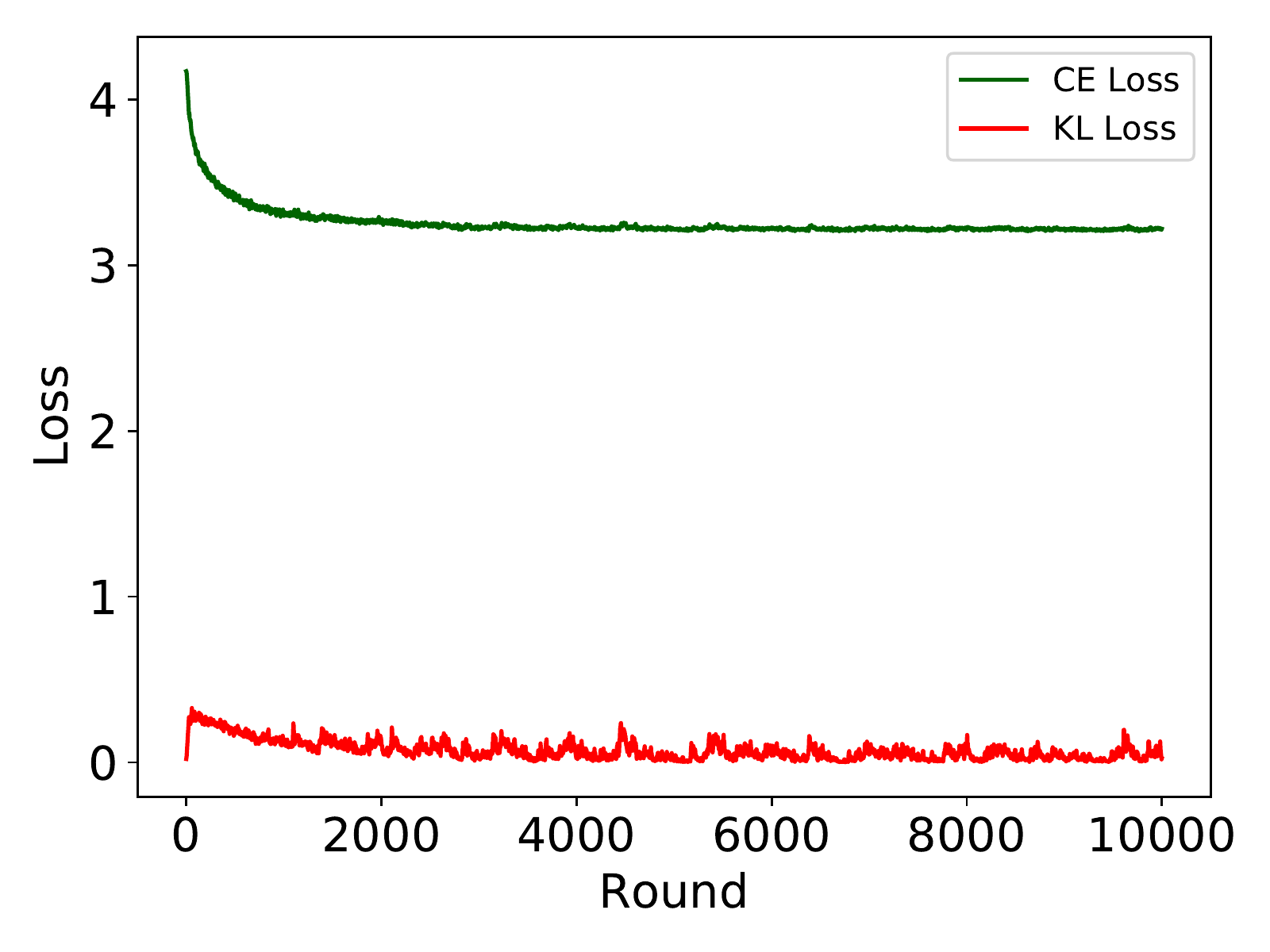}
\caption{Learning and loss curves on Office-Home in domain Product. Left: ACC on validation data. Middle: Loss in local optimization. Right: CE and KL losses in global optimization of \modelName{}.} \label{fig:curve}
\end{figure}

\keypoint{Optimization and loss analysis} \cut{We provide further analysis of \modelName{} in Figure~\ref{fig:curve}. Taking the learning of AlexNet in domain Product of Office-Home as an example,} Figure~\ref{fig:curve}(left) shows ACC on the validation data. \modelName{} exhibits faster convergence to the higher performance. Figure~\ref{fig:curve}(right) shows the consistent utility of KL loss during the first 1000 rounds for convergence and performance benefits as shown on ACC. Figure~\ref{fig:curve}(middle) shows the loss during local optimization, which benefits \modelName{} locally with the help of the global mutual learning. \cut{We further discuss KL and CE losses during  global optimization in details in the follow-up ablation study.}


\begin{table}[!htbp]
  \centering  
  \fontsize{8.4}{8}\selectfont  
  \caption{Components study in global mutual optimization (Avg).}  
  \label{tab:ablation}  
    \begin{tabular}{lccc}  
    \toprule  
    Method & ACC&WDP&CDP\cr  
    \midrule  
    \modelName{} & {\bf89.13}&{\bf93.33}&{\bf87.72}\cr
    \modelName{} (no KL) & 86.79&91.67&84.50\cr
    \modelName{} (no $project$)& 88.46&92.67&87.45\cr 
    \modelName{} (PCGrad)&88.34&92.67&86.89\cr
    \bottomrule  
    \end{tabular}  
\end{table}  

      

\keypoint{Ablation on design components of global mutual optimization}
In global optimization, our contributions are: KL mimicry loss Eq.~(\ref{eq:losskl}), and the $project$ operation for the calculation of $\tilde{g}_i$ to achieve stable multi-domain learning Eq.~(\ref{eq:gemprimal}). We ablate them in Table~\ref{tab:ablation} on Rotated MNIST ($\alpha=10\%$).

KL loss plays an important role in both CDP and WDP. The robustness benefit of mutual learning by KL loss to find a \textit{wider} minimum in the single domain has been analyzed in DML \citep{zhang2018dml}. Similarly, under our multi-domain setting,\cut{ KL loss guides to match the current network's predictions to match the other teacher networks'  posterior predictions based on the corresponding node/domain's public data, which} the matching with teachers' posterior predictions increases the model's generalization (CDP) to other domains. Meanwhile, the soft labels (for KL loss) help to alleviate the domain shift interference of the domain's hard true labels (for CE loss). Thus KL loss benefits optimization stability (WDP) during the global mutual optimization. 

If we remove the $project$ operation, then $\theta_i$ will be updated by directly using $g^{\text{pub}}_{i}$\cut{ without respecting the constraint in Eq.~(\ref{eq:constraints})}. 
The results confirm that WDP gets worse without the constrained $\tilde{g}_i$. Moreover, we compare with an alternative gradient projection \cut{method }PCGrad \citep{yu2020gradient} which deals with conflicting gradients in a handcrafted way\cut{ rather than being QP-based in \modelName{}}. But PCGrad shows unsatisfactory performance even slightly worse than without the project operation. 

\cut{How does MAFML help to find such a better ``plasticity'' without sacrificing ``stability'' from other domains through the global mutual optimization? In the process of ``teaching each other'' on global mutual gradient updates, besides learning to match the probability estimate of the ensemble domains in Eq.~(\ref{eq:losskl}) which have high-entropy posterior leads to more robust
KL update, we also obtain CE update through the CE loss Eq.~(\ref{eq:ce_pub}) on other domains public data which has optimal non-label option. Moreover, in order to get stability, we achieve the constraint in Eq.~(\ref{eq:constraints}) with inner dot update. To gain a better understanding of the contribution of each component  to  the  final model, we observe the performance changes in Rotated MNIST when the components of gradient update is not applied.}

\keypoint{Hyperparameter sensitivity}
We ablate the hyperparameter of $E$ in \modelName{} in Table~\ref{tab:tabE} on Rotated MNIST ($\alpha=10\%$). \cut{The results considering varying $E$ on PACS and Office-Home are reported in the supplementary materials.} \modelName{} generally performs better with lower update interval $E$. Performance degrades smoothly with larger $E$ which lowers communication cost proportionally.

\begin{table}[!htbp]
  \centering  
  \fontsize{8.4}{8}\selectfont  
  \caption{Hyperparameter sensitivity of $E$ in \modelName{} (Avg).}  
  \label{tab:tabE}  
    \begin{tabular}{lccc}  
    \toprule  
    Method & ACC&WDP&CDP\cr  
    \midrule  
    \modelName{} ($E=1$) & {\bf89.13}&{\bf93.33}&{\bf87.72}\cr
    \modelName{} ($E=5$) & 88.04&92.17&86.67\cr
    \modelName{} ($E=10$)& 87.25&93.17&85.28\cr 

    \bottomrule  
    \end{tabular}  
\end{table}  



\keypoint{Limitations} A limitation of \modelName{} is while our comms cost is $\approx10e6\times$ lower than FedAvg at small scale (4 nodes), this advantage will be eroded if scaled to many participants. This could be alleviated by communicating between a subset of randomly chosen pairs at each global round, which preliminary experiments of such asynchronous distributed learning in Table~\ref{tab:performance_comparison_mnist} show lead to similar performance.

\cut{

\begin{table}[!htbp]
  \centering  
  \fontsize{8.4}{8}\selectfont  
  \caption{Synchronous vs. asynchronous \modelName{} ($\alpha=10\%$) using Rotated MNIST (Avg).}  
  \label{tab:asynchronous}  
    \begin{tabular}{lccc}  
    \toprule  
    Method & ACC&WDP&CDP\cr  
    \midrule  
    \modelName{} (synchronous) & {\bf89.13}&{\bf93.33}&{\bf87.72}\cr
    \modelName{} (asynchronous) & 88.21&92.17&86.89\cr

    \bottomrule  
    \end{tabular}  
\end{table} 

}

\section{Conclusion}
We proposed \modelName{} for FL with heterogeneous models and data statistics. Each node in the cohort acts as both student and teacher, providing effective communication efficient federated learning. \cut{Crucially, }\modelName{} supports heterogeneous architectures, which is crucial for FL across diverse hardware platforms, and with institutions' proprietary models; and is robust to heterogeneous data statistics, which -- while not widely studied academically -- is ubiquitous in practical FL\cut{ scenarios}. 



\bibliographystyle{named}
\bibliography{ijcai21}

\end{document}